\documentclass[letterpaper, 10 pt, conference]{ieeeconf}  

\IEEEoverridecommandlockouts                              

\usepackage{subfigure}
\usepackage{overpic}
\usepackage{times}
\usepackage{epsfig}
\usepackage{graphicx}
\usepackage{amsmath}
\usepackage{amssymb}
\usepackage[]{algorithm2e}
\usepackage{xcolor}
\usepackage{hyperref}

\newcommand{\DLO}{\mbox{DeLiO}}

\title{\LARGE \bf
\DLO{}: Decoupled LiDAR Odometry
}

\author{Queens Maria Thomas$^{1}$  \quad Oliver Wasenm\"uller$^{1}$ \quad Didier Stricker$^{1}$
\thanks{$^{1}$German Research Center for Artificial Intelligence - DFKI,
        Kaiserslautern, Germany,
        {\tt\small firstname.lastname@dfki.de}}%
}

\begin{document}

\maketitle
\thispagestyle{empty}
\pagestyle{empty}

\begin{abstract}
Most LiDAR odometry algorithms estimate the transformation between two consecutive frames by estimating the rotation and translation in an intervening fashion. 
In this paper, we propose our Decoupled LiDAR Odometry (\DLO{}), which -- for the first time -- decouples the rotation estimation completely from the translation estimation. 
In particular, the rotation is estimated by extracting the surface normals from the input point clouds and tracking their characteristic pattern on a unit sphere. 
Using this rotation the point clouds are unrotated so that the underlying transformation is pure translation, which can be easily estimated using a line cloud approach. 
An evaluation is performed on the KITTI dataset and the results are compared against state-of-the-art algorithms.
\end{abstract}

\section{Introduction}

Odometry plays an essential role in many applications such as robot navigation or autonomous driving where robust and precise estimation of the trajectory is a main challenge. 
Nowadays, this has become more interesting as a variety of sensors that could include the spatial information as well have emerged \cite{yoshida2017time}. 
A LiDAR (Light Detection And Ranging) is such an active optical sensor that can provide highly accurate range measurements where errors are usually constant irrespective of the distance. 

Odometry is often estimated by incrementally integrating the relative transformations resulting from consecutive point cloud registrations. 
Traditional point cloud registration approaches such as Iterative Closest Point \cite{besl1992method,icp} alternate between the translation and rotation estimation in an iterative fashion. 
This has the overhead of estimating the three degrees of translational freedom and three degrees of rotational freedom simultaneously. 
In this paper, we propose the new \DLO{} algorithm that estimates the transformation by decoupling the rotation from translation and thus making the six DoF odometry estimation boil down to two times three DOF.
To the best of our knowledge, we are the first to propose this decoupling for LiDAR odometry.

\begin{figure}[t]
\centering
\includegraphics[width=0.65\linewidth]{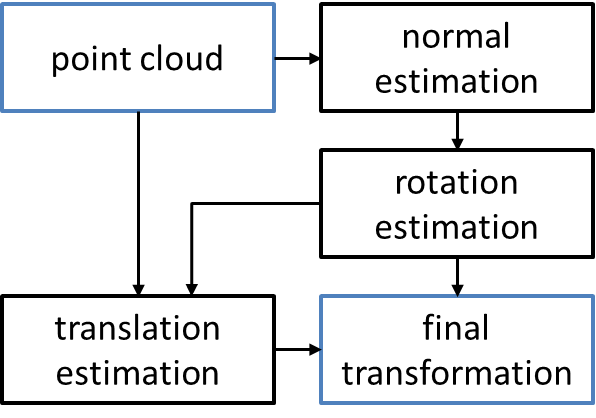}
\caption{
Overview of \DLO{}:
Normals are extracted out of the input point clouds and rotation is estimated independently of translation.
Translation is estimated out of the unrotated point clouds.
}
\label{fig:overview}
\end{figure}

The decoupling of rotation estimation in \DLO{} is driven by the following main observation in two consecutive LiDAR point clouds:
The rotational difference between surface normals corresponds to the rotation difference between their corresponding LiDAR scans.
This property is independent of any translation, since translation does not affect the surface normals.
Thus, this can be exploited to estimate the rotation independent of translation.
The main challenge in \DLO{} is to determine the rotation difference in the normals of two LiDAR scans.
In Section \ref{sec:rotation} we propose a new approach to track these surface normals over time.
Later, point clouds are unrotated using the estimated rotation so that the underlying transformation is a pure translation. 
This translation is estimated in Section \ref{sec:translation} using a state-of-the-art line cloud odometry inspired from the work of Velas et al. \cite{cls}. 
In Section \ref{sec:evaluation} we verify the state-of-the-art performance of  \DLO{} on the KITTI benchmark.
An overview of our new approach is given in Figure \ref{fig:overview}.

\begin{figure}[t]
\subfigure[LiDAR scanner]{\includegraphics[width=0.23\textwidth]{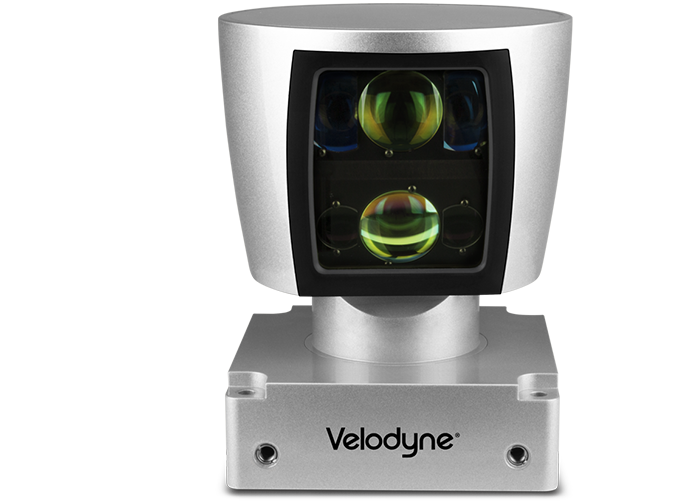}} \hfill
\subfigure[LiDAR point cloud]{\includegraphics[width=0.23\textwidth]{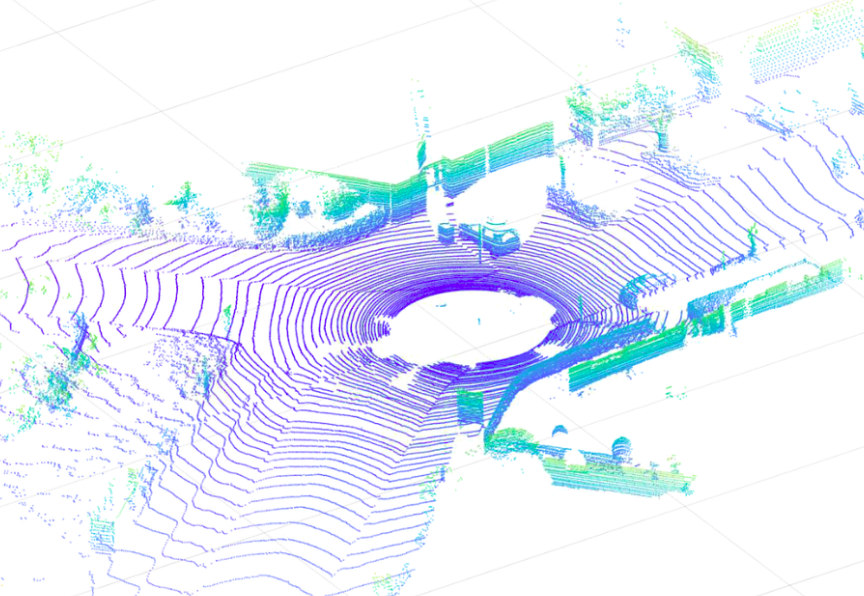}}
\subfigure[Normals on unit sphere]{\includegraphics[width=0.24\textwidth]{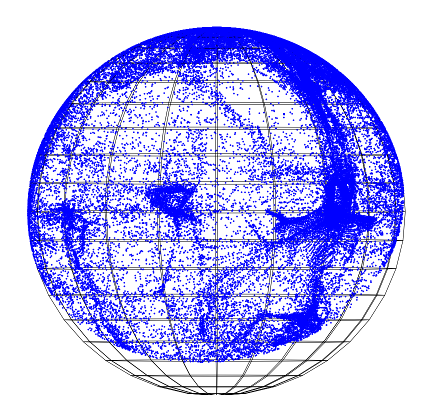}} \hfill
\subfigure[\DLO{} trajectory]{\includegraphics[width=0.23\textwidth]{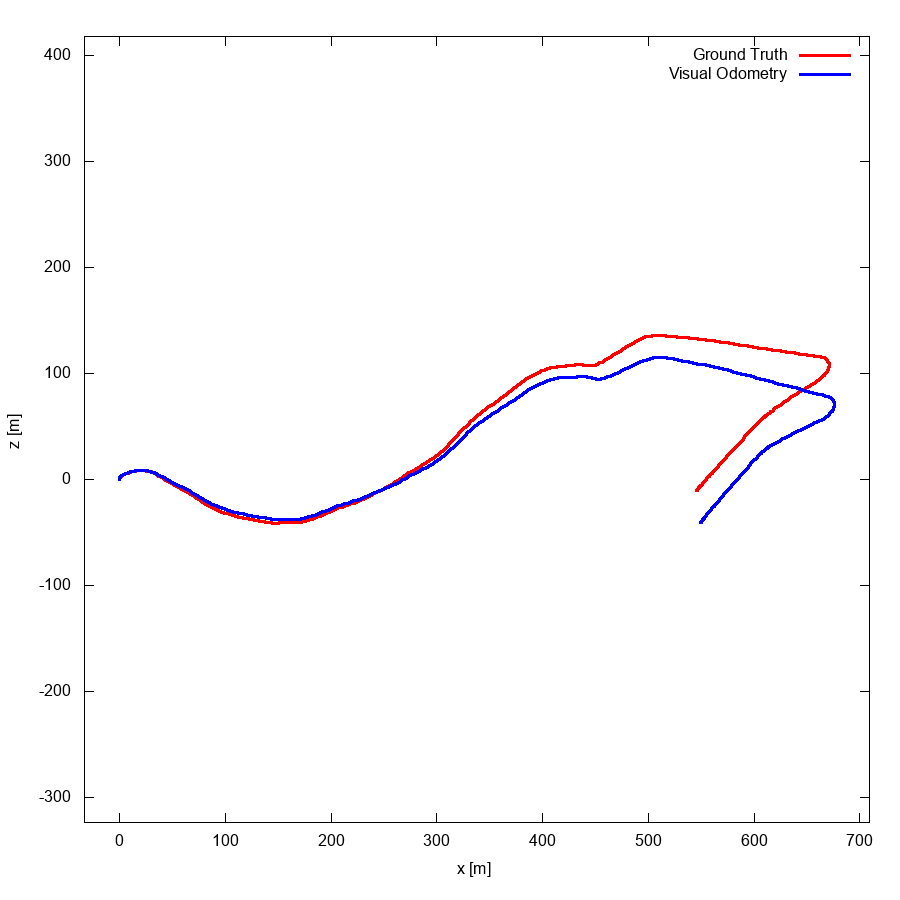}}
\caption{
In this paper, we propose \DLO{} -- a LiDAR odometry decoupling rotation from translation estimation.
Decoupling is achieved by tracking surface normals on a unit sphere for rotation estimation, which is independent of translation estimation.
}
\label{fig:teaser}
\end{figure}


\section{Related Work}

In the literature plenty of odometry algorithms exist for several visual sensor types \cite{rel_1}. 
Several algorithms -- such as LIMO \cite{graeter2018limo}, DEMO \cite{zhang2014real}, etc. -- estimate the odometry out of a monocular or stereo camera and use the LiDAR points for support only.
However, for our approach we want to rely on pure LiDAR information.

Since LiDAR information is a large-scale point cloud, point cloud registration is usually applied in the state-of-the-art for LiDAR-only odometry.
For point cloud registration often variations of the traditional Iterative Closest Point (ICP) \cite{icp} are utilized. 
ICP and its variations estimate the transformation involved between two scans by iteratively estimating rotation and translation in an alternating fashion. 
But, ICP is prone to local minima when the point clouds are considerably far from each other. 
This is because the point cloud correspondences can be accurately estimated only when initial alignment is already very good. 
Also ICP is highly sensitive to noise and outliers. 
The exact point to point matching in standard ICP does not take into account the differential sampling of the two point clouds and thus may lead to pairs without good equivalence. 
A few ICP variants were proposed to overcome this issue by taking information from surface normals into consideration. 
Two prominent variants are \textit{Point-To-Plane} approach that considers the surface normals from the target point cloud only and \textit{Plane-To-Plane} variant that uses the normals from both source and target point cloud.

A recent state-of-the-art approach in odometry using LiDAR sensors is LiDAR Odometry and Mapping in Real-time (LOAM) \cite{loam} that extracts features that are on sharp edges and planar surface patches using the scan sweep information. 
Also, LOAM integrates data from IMU units as well to improve the accuracy of estimation. 
Visual-LOAM (V-LOAM) \cite{vloam}, is a variation of LOAM that aims at motion estimation and mapping using a monocular camera with a 3D LiDAR instead of IMU. 
A recent variation of ICP is Implicit Moving Least Squares-SLAM (IMLS) \cite{imls}, that relies on a scan-to-model matching framework. 
Initially the LiDAR scans are sampled with a sampling strategy and then a model is defined from the previous localized LiDAR sweeps and use an Implicit Moving Least Squares (IMLS) surface representation.
Unlike LOAM, IMLS does not use any data from other sensors like IMU or cameras but removes dynamic objects from the scene as a pre-processing step. 
The DLO \cite{sun2018dlo} approach estimates the odometry directly, by projection the LiDAR scan to a ground plane and registering the resulting grid maps.

Unlike the before mentioned approaches, our method \DLO{} decouples for the first time the rotation completely from translation in LiDAR odometry. 
Thus, rotation can be estimated in one step.
The point clouds are later unrotated with the estimated rotation so that the underlying transformation is pure translation, which is estimated using a modified line cloud odometry. 
A further refinement of the rotation is no longer necessary in comparison to the state-of-the-art algorithms.
Also \DLO{} does not use any data from other sensors like IMU, cameras or GPS.

\section{Rotation Estimation}
\label{sec:rotation}

\DLO{} determines odometry by decoupling rotation from translation estimation.
Thus, a robust rotation estimation scheme is necessary, which can be estimated independently of the translation.
Utilizing the 3D coordinates captured by the LiDAR scanner directly is not feasible, since their alignment involves rotation as well as translation.

\begin{figure*}
\begin{center}
 \includegraphics[width=0.23\linewidth]{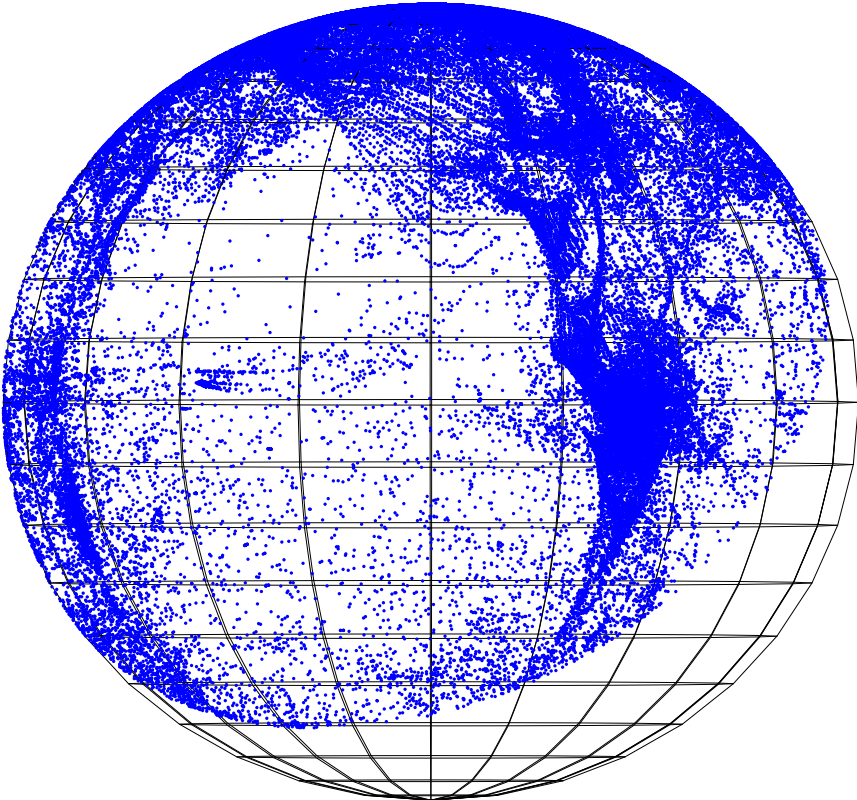}
 \includegraphics[width=0.23\linewidth]{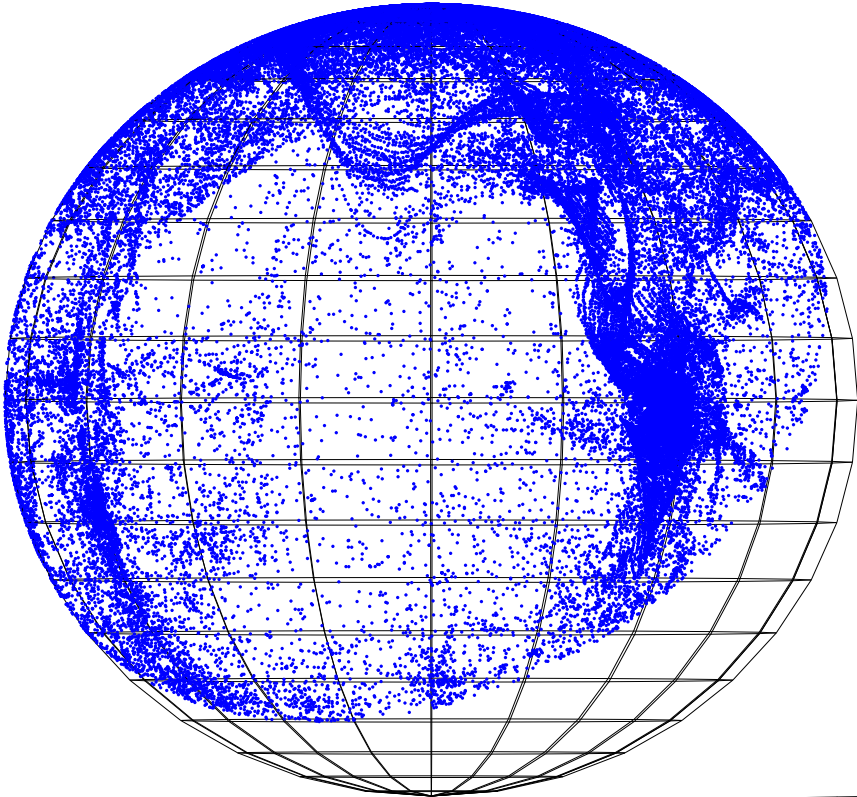}
 \includegraphics[width=0.23\linewidth]{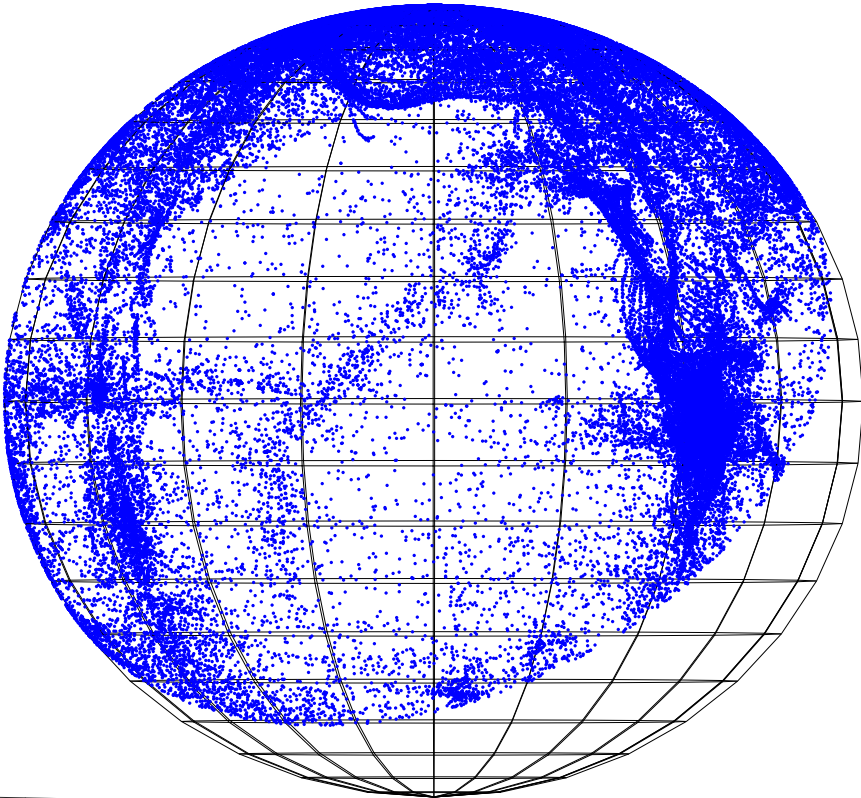}
 \includegraphics[width=0.23\linewidth]{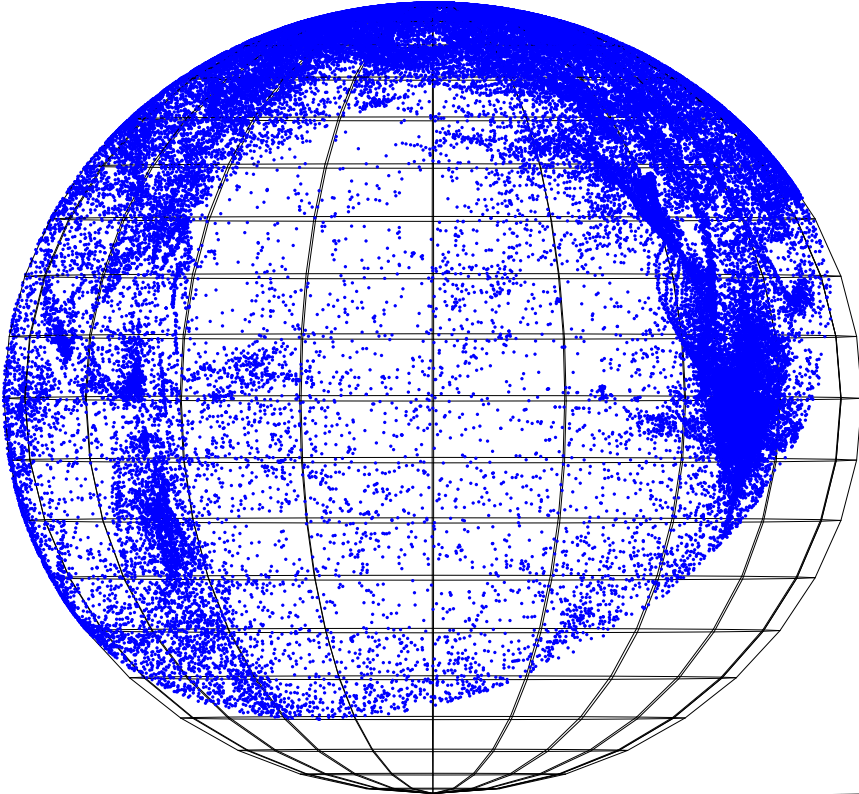}
\end{center}
   \caption{Surface normal mapping on the unit sphere for four consecutive frames. A characteristic pattern is moving over the unit sphere over time. This rotational movement corresponds to the rotation between the corresponding LiDAR scans, independent of the translation.}
\label{fig:pattern}
\end{figure*}

Therefore, our main idea is to use the surface normals of the captured LiDAR scans for rotation estimation.
Surface normals can  be  extracted  directly  from  the captured scans  and  convey rich  geometric  information.
The normal vector distributions typically contain a characteristic structure depending on the captured environment as visible in Figure \ref{fig:pattern}.
These structural regularities lead to robust patterns in the normal vector density distribution.
The rotational difference between two subsequent patterns corresponds to the rotation difference between their corresponding LiDAR scans.
Thus, we track the normal distribution over time in order to determine rotation.
This can be done completely independent of the translation difference.

In the following, our new rotation approach for \DLO{} is described in more detail.
We start with the normal extraction out of the captured LiDAR scans.
In order to distinguish characteristic patterns in the normal distribution from random noise, we apply a statistical outlier filtering. 
If the retained subset is not enough for a pattern registration, a mean shift seeking is performed to extract more patterns from the surface normals. 
Later rotation is estimated out of the displacement of these patterns using a Singular Value Decomposition (SVD).

\subsection{Surface Normal Extraction}

Surface normals are extracted, since they are translation independent geometric properties. 
A least-square plane fitting estimation problem \cite{pcl} 
that estimates the normal of a point on a surface by approximating the normal of a plane tangent to the surface is used for surface normal estimation. 
The analysis of eigenvectors and eigenvalues (estimated using Principal Component Analysis (PCA) \cite{pca}) of a covariance matrix formulated from the k-nearest neighbors \cite{knn_paper} of each query point helps in estimating the surface normals corresponding to the query points.
The covariance matrix is formulated as
\begin{equation}
\mathcal{C} = \frac{1}{k} \sum_{i=1}^{k}{\cdot (\boldsymbol{p}_i-\overline{\boldsymbol{p}})\cdot(\boldsymbol{p}_i-\overline{\boldsymbol{p}})^{T}}, ~\mathcal{C} \cdot \vec{{\mathsf v}_j} = \lambda_j \cdot \vec{{\mathsf v}_j},
\end{equation}
where $k$ is the number of nearest neighbors of point $p_i$, $\overline{p}$ is the centroid of the $k$ nearest neighbors of $p_i$,  and $\vec{{\mathsf v}_j}$ is the $j$-th eigenvector of the covariance matrix, and $\lambda_j$ is the $j$-th eigenvalue.
In order to be more robust against outliers, bilateral filtering on the surface normals is applied \cite{wasenmuller2015combined}.
 
Prominent aspects from the surface normals are now identified by analyzing the pattern formed by these normals on an unit sphere. 
The motivation behind this idea is the fact that the resulting characteristic pattern on unit sphere is moving over the unit sphere over time as depicted in Figure \ref{fig:pattern}. 
Tracking this pattern over time yields rotation independent of translation.


\subsection{Statistical Outlier Removal}
\label{sec:statistical}

The presence of large number of scattered points on the unit sphere affects adversely the tracking of the pattern. 
Hence a robust outlier removal scheme has to be considered to select a subset of the points on the unit sphere that effectively represent the underlying pattern. 

\begin{figure*}
\begin{center}
 \includegraphics[width=0.23\linewidth]{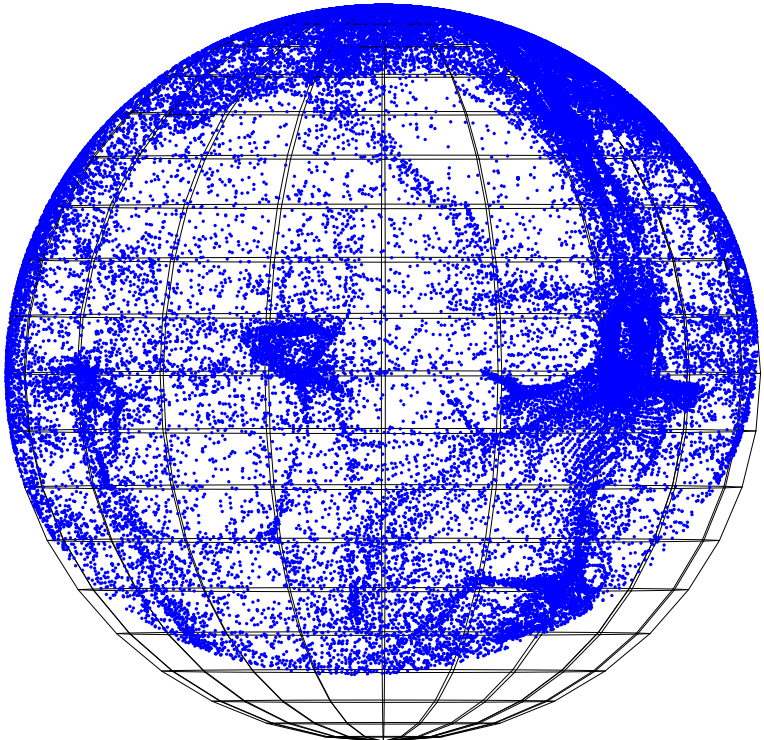}
 \includegraphics[width=0.23\linewidth]{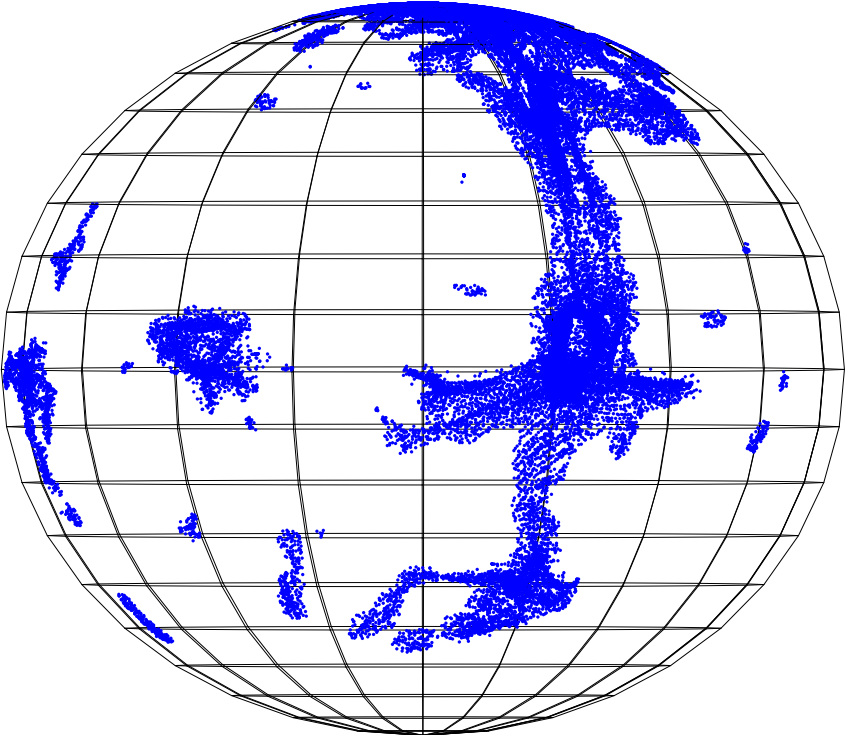}
 \includegraphics[width=0.23\linewidth]{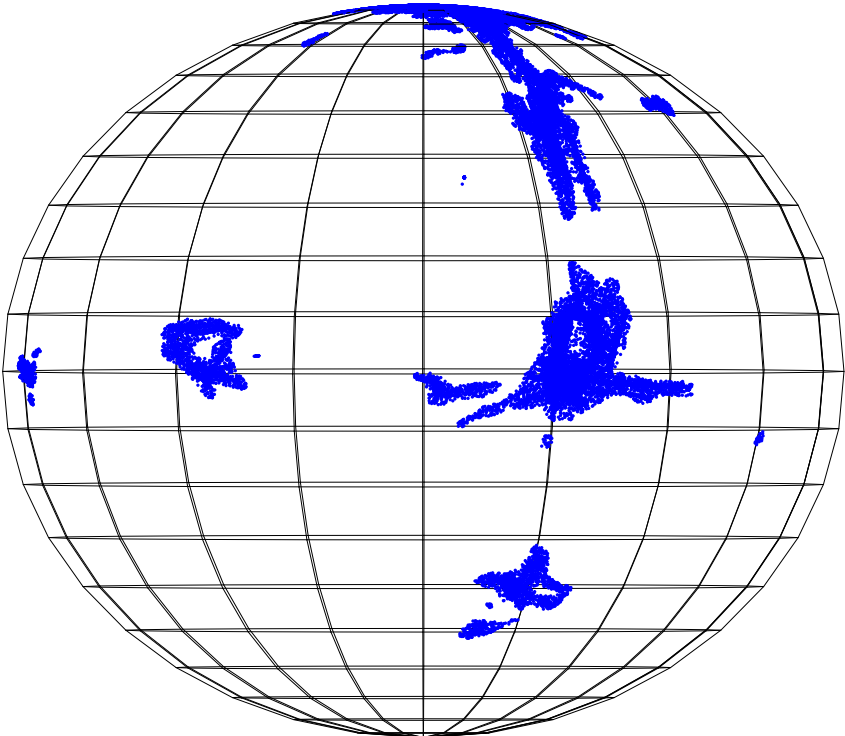}
 \includegraphics[width=0.23\linewidth]{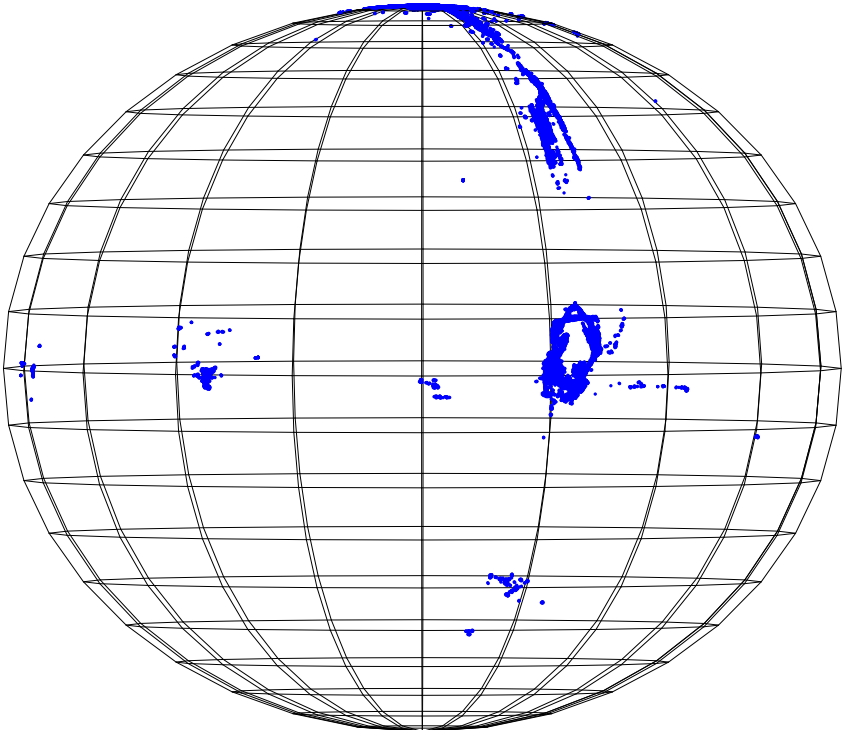}
\end{center}
   \caption{Statistical Outlier Removal (SOR) applied iteratively (from left to right) on the unit sphere of surface normals.}
\label{fig:sor}
\end{figure*}

Therefore, we utilize a statistical outlier removal scheme \cite{statistical_outlier} that performs a statistical analysis in the neighborhood of each point, removing those points violating a certain criteria. 
For every point the mean distance to all its neighbors are calculated. 
Under the assumption that the resulting distribution is Gaussian with a mean and standard deviation, all points whose mean distances are outside a defined interval are considered as outliers and removed. 
Considering various number of neighboring points, this is done iteratively to retain a small but stable set of points. 
The resulting surface normals on the unit sphere are shown in Figure ~\ref{fig:sor}.

If the scene lacks significant characteristic features, the resulting pattern on the unit sphere may be sparse and does not convey meaningful information to estimate the rotation. 
If the number of points retained after statistical outlier removal is lower than a threshold value, the may mean that the only remaining points represent ground because of the large number of ground points resulting in very dense areas over the unit sphere. 
A mean shift seeking \cite{MeanShiftAlgo} explained in detail in the following section is used if the retained number of points fall below a certain threshold value. 
This would facilitate finding more structures or dense points corresponding to new structures other than ground. 
Once the dense positions are discovered, a ball query discovers neighbors in a certain radius around each of the dense points. 
Since these points are prone to outliers or noise, statistical outlier removal is performed until a good number of points remain. 
The obtained points are downsampled and filtered using uniform sampling. 
Here a 3D voxel grid is assembled over the input points and in each voxel, all the points are assumed with their centroid. 
This approach reduces the number of points significantly, retaining only the best points and no outliers.

%
%

\subsection{Mean Shift Seeking}

As explained before, in case too few normals are retained after the Statistical Outlier Removal, a mean shift seeking is applied in order to preserve also fine structures for normal tracking.
The mean shift seeking procedure applies local Kernel Density Estimation (KDE), if an initial approximate location of the mode is known and takes steps iteratively in the direction of increasing density. 
To ensure that the whole surface of the sphere is covered, the sphere surface is divided into a grid of uniform size which gives a set of coordinates that can be used to initialize the mode seeking. 
A nearest neighbor approach was used to clean the identified set of modes since mean shift iterations may converge to same modes if the initialization points are close to each other. 
The idea is to perform a single mean shift iteration for each mode given a subset of normal vectors on unit sphere. 
The subset is selected by considering all normal vectors that are in the neighborhood of considered center $f_j$. 
The range of the neighborhood is a conical window centered at the center of the sphere with an apex angle of $\theta_{window}$. 
The required subset of normal vectors $n_i$ lie within this conical window and satisfies the condition 
\begin{equation}
\|n_i \times f_j\| < sin(\theta_{window}/2).
\end{equation}

To compute the mean shift, these normal vectors are then projected into the tangential plane at $f_j$. 
A Riemann exponential map \cite{MMF} is applied so that the projected vectors represent proper angular deviations in the tangential plane. 
Then the mean shift in the tangential plane is computed as
\begin{equation}
   S'_j = \frac{\sum_{ij} e^{-c\|m'_{ij}\|^2}m'_{ij}}{\sum_{ij} e^{-c\|m'_{ij}\|^2}},
\end{equation}   
where $m'_{ij}$ are the tangential plane coordinates. 
$c$ is a design parameter that defines the width of the kernel in the tangential plane and can be derived from $\theta_{window}$. 
The mean shift after computation is transformed onto the unit sphere again using Riemann logarithmic map.


\subsection{Pattern Registration}

Finally, rotation is estimated using a Singular Value Decomposition (SVD) \cite{SVD}. 
For each target point, the corresponding source point is obtained using a k-nearest neighbor method. 
A KD-Tree \cite{knn_paper} is constructed using the source point set and thus reducing the search space significantly. 
Once the correspondences are obtained, SVD can be used to estimate rotation. 
This is performed iteratively till estimated rotation converges.
Using our new described approach makes it possible to estimate rotation completely independent from translation.
Also an alternation between rotation and translation estimation in not necessary, since the final rotation is estimated after executing the procedure explained in this Section.


\section{Translation Estimation}
\label{sec:translation}

After the rotation is estimated, this result can be used to unrotate the two LiDAR scans.
The remaining transformation between them is then a pure translation.
A major problem involved in LiDAR point cloud registration is the sparsity of the point cloud. 
Also the ring structure of the point cloud leads to incorrect registration \cite{cls}. 
Since classical ICP approaches \cite{besl1992method,icp} try to minimize the distance of closest points, the resulting transformation could be the one that fits the ring structures that represent the floor because of the larger number of floor ring points. 
These issues associated with point clouds can be effectively handled by a line cloud proposed by Velas et al. \cite{cls}. 
In contrast to their algorithm where rotation and translation are estimated simultaneously, we use this approach for translation estimation only.

The line cloud approach has two main parts. 
First, the points in the point cloud are transformed into polar coordinate system and line cloud is generated. 
Second, the generated lines are registered and translation is obtained by aligning the centroids.
Conversion to line cloud addresses the sparsity issue of the point cloud by generating lines such that the lines represent the underlying scene geometry efficiently. 
The original point cloud is iteratively resampled so that the new points for registration do not come from the point cloud, but the closest points between two matching pair of lines.


\subsection{Line Cloud Generation}

As an initial step, each 3D point is converted from Cartesian coordinate system to a polar coordinate system for the line cloud generation. A polar coordinate system is a two dimensional coordinate system where each point is determined by a distance, \textit{r} from the origin (radial coordinate) and an angle, \textit{$\theta$} from the reference direction (angular coordinate). The Cartesian coordinates [\textit{x,y,z}] of the point cloud can be converted to polar coordinates \textit{r} and \textit{$\theta$} by
\begin{equation}
r={\sqrt {x^{2}+y^{2}}}\quad
\theta =\operatorname {atan2} (y,x)\quad \theta\in[0,360),
\end{equation}
where $\theta$ is the angle within ground plane \textit{x-y} and \textit{r} represents the distance from origin. Vertical axis \textit{z} is not considered for the conversion because of the horizontal ring layout of the point cloud.
\begin{figure}[t]
\begin{overpic}[width=\linewidth]{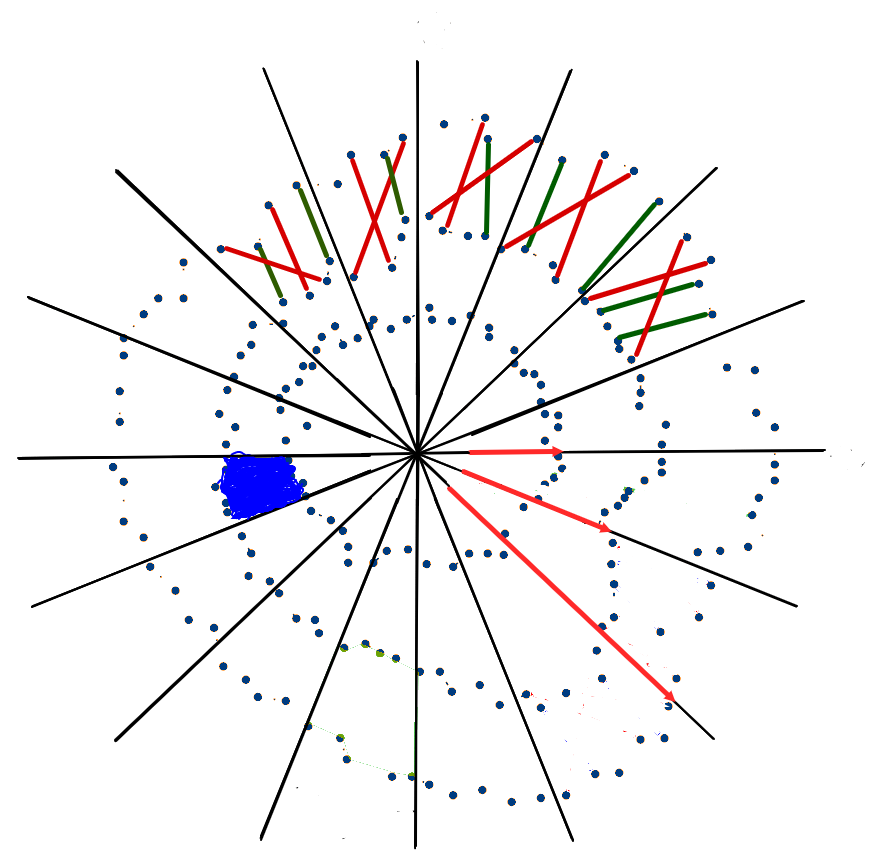}
       \put(58,49){\makebox(2,2)\textbf{\color{red}1}}
       \put(64,41){\makebox(2,2)\textbf{\color{red}2}}
       \put(73,22){\makebox(2,2)\textbf{\color{red}3}}
       \put(55,44){\makebox(2,2)\textbf{\color{red}$\theta$}}
    \end{overpic}
\caption{Line Cloud generation \cite{cls}: 
The area around the LiDAR scanner is divided into rings and bins so that the points are represented in polar coordinate system. 
One sample bin is shown in blue color. 
Lines (red and green) are drawn connecting points in consecutive rings and bins. 
Only shortest lines (green) are preserved. 
}
\label{fig:line_cloud_regis}
\end{figure}
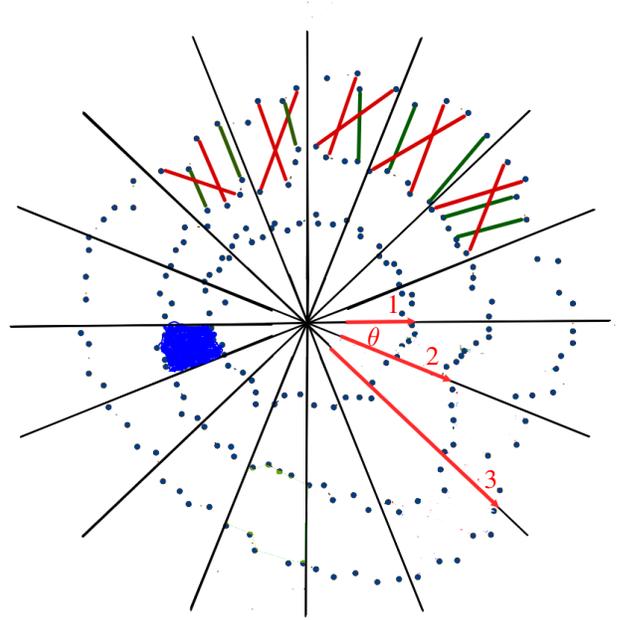

The LiDAR scanner produces rays such that each ray captures a ring of points as depicted by the red arrows in Figure \ref{fig:line_cloud_regis}.
Points in the point cloud are shown as blue dots. 
Line cloud is generated by connecting some of the promising points in the point cloud in such a way that the underlying local surface properties are captured. 
The scanner position is considered as origin and the space around is divided into polar bins. 
Each point in the point cloud is represented by the ring and bin. 
One sample bin is shown by blue color in Figure \ref{fig:line_cloud_regis}. 
To ensure points of similar angle are selected, random lines are generated in each bin connecting points from consequent rings. 
Only few  promising points from the line cloud are selected using a filtering criteria. 
Preserving only shorter lines (green lines) ensures that real planes are connected and points connecting different planes (red lines) are discarded.


\subsection{Line Cloud Registration}
Once the point cloud is sampled into line cloud, the translation is estimated using a variation of ICP. Since the point clouds are unrotated with absolute rotation, translation is calculated iteratively by finding closest matches from the line cloud. Instead of finding point-to-point correspondences, the algorithm finds correspondences between source and target line segments. This is done by finding the line segment in the target cloud whose center is close to the center of source line segment using a KD-Tree implementation. A distance threshold is included to eliminate incorrect matches in such a way that only the source and target line center pairs whose distance is less than the mean distance of all match pairs  are kept.

Once the matching pair of line segments are found out, the line segments are aligned into a 3D plane that gives the translation involved. Consider a matching pair of lines \textit{$l_s$} and \textit{$l_t$} that are taken from the source cloud and target cloud respectively. The parametric representation of the lines given 3D points $P_{s0}$ and $P_{t0}$ and vectors $u_s$ and $u_t$ are
\begin{equation}
l_s : P_s = P_{s0} + t_s\cdot{u_s};\qquad t_s\in(-\infty,\ \infty)
\end{equation}
\begin{equation}
l_t : P_t = P_{t0} + t_t\cdot{u_t};\qquad t_t\in(-\infty,\ \infty),
\end{equation}
where $u_s$ and $u_t$ are vectors representing the direction of the line segments. 

In any $n$ dimensional space, the two lines $l_s$ and $l_t$ are closest at points $X_s$ and $X_t$ when $w$ is the unique minimum of $w(P_s,P_t)$ \cite{math_guide}. These closest points \textit{$X_s$} and \textit{$X_t$} are the correspondences between the source and target line clouds. The closest points between these lines are given by
\begin{equation}
X_s = P_{s0} + t_{sc}\cdot{u_s}\quad
\end{equation}
\begin{equation}
X_t = P_{t0} + t_{tc}\cdot{u_t}\quad,
\end{equation}
where, 
\begin{equation}
    t_{sc}=\frac{be-cd}{ac-b^{2}}\qquad t_{tc}=\frac{ae-bd}{ac-b^{2}}
\end{equation}
\begin{equation}
    a = {u_s}\cdot {u_s}\qquad b = {u_s}\cdot {u_t}\qquad
    c = {u_t}\cdot {u_t}
\end{equation}
\begin{equation}
    d = {u_s}\cdot {w}\qquad
    e = {u_t}\cdot {w}\qquad   w = P_{s0} - P_{t0}
\end{equation}

The final transformation is estimated from the centroids of both sets of points under the assumption that the underlying transformation is pure translation. 
When the line clouds are registered, the distance between the preserved lines become very small or close to zero and the lines are coplanar.


\section{Optimization}
\label{sec:optimization}

Odometry integrates small relative transformations incrementally over time. 
The incremental nature of this estimation is subject to drift that accumulates to large errors \cite{drift} 
This drift can be reduced to a certain extend using a variety of optimization techniques. 
Estimated transformations can be improved using multi-frame processing \cite{multiframe} as well as prediction based on constant velocity motion model.


Since \DLO{} relies on a tracking of the surface normals for rotation estimation, a good initialization is beneficial.
The closer the prediction is to the actual rotation value, the better is the final rotation result.
Thus, we use a linear prediction model. 
The prediction $T_{p}$ for pairs of frames at time $i$ and $i+1$ can be estimated from previous $N$ frames using the linear prediction as
\begin{equation}
    T_{p} = \frac{2}{N(N+1)}\sum_{j=1}^N{(N-j+1)T_{i-j}}.
    \label{eq:linearPrediction}
\end{equation}
%
%



Even though prediction improves results considerably, there is increased accumulation of drift as the length of trajectory increases. 
A recent trend is to use graphical models \cite{graphOptim} to solve the accumulated drift through optimization. 
One way of formulating a pose-graph for optimization is to construct a graph whose nodes correspond to the poses of the sensor at different points in time and whose edges represent constraints between the poses. 
The constructed pose-graph can be further optimized by minimizing a non-linear error function that corresponds to the graph. 

Graph based optimization approaches build a pose graph incrementally taking the sensor pose at each point of time. 
The first node is constructed using the initial pose of the sensor. 
As soon as the sensor moves to a new position, the transformation between these positions is computed using the \DLO{} algorithm of Sections \ref{sec:rotation} and \ref{sec:translation}. 
When this transformation is applied to position $a$, a new measurement at position $a+1$ is acquired. 
The edge between positions $a$ and $a+1$ correspond to this estimated transformation. 
When the car moves further, a new transformation is available which when applied to position $a+1$ gives measurement at position $a+2$. 
The transformation between position $a$ and $a+2$ can be estimated if a sufficient overlap exists between these two positions. 
A third edge may be added between position $a$ and $a+2$ in the pose-graph. 
Due to imperfections in the incremental registration and noise from the environment, the two estimations at position $a+2$ may differ a bit. 
This error can be minimized by fine tuning the graph using optimization algorithms that update the pose with the optimized values.

Before optimization, an error function needs to be defined. 
A non linear least squares optimization can be described by
\begin{equation}
     {F}({ x})=\sum_{\langle k\rangle\in {\cal C}}\underbrace{{ e}({ x}_{k},{ z}_{k})^{\top}\Omega_{k}{e}({ x}_{k},{ z}_{k})}_{{ F}_{k}}
\end{equation}
\begin{equation}
    {x}^{\ast} = \mathop{\rm argmin}_{{x}}{ F}({x}),
\end{equation}
where $x$ is a vector of parameters. 
$z_k$ and $\Omega_k$ represent the mean and information matrix relating the parameters in $x_k$. 
$e(x_k,z_k)$ defines how well the parameters $x_k$ fits the constraints in $z_k$ and $\Omega_k$. 
The information matrix contains value indicating how good the current measurement is. 
This is the inverse of covariance matrix. The Levenberg-Marquardt algorithm \cite{levenberg} implemented in $g^2o$ \cite{g2o} is used to carry out the optimizations. 


\section{Evaluation}
\label{sec:evaluation}


\begin{figure}[t]
\includegraphics[width=\linewidth]{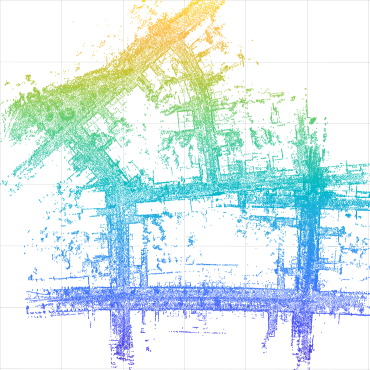}
\caption{ 
Example of environment map (KITTI \textit{training} sequence \#07) utilizing the \DLO{} trajectory. 
}
\label{fig:pointcloud}
\end{figure}

\begin{figure}[t]
\begin{center}
 \includegraphics[width=0.45\linewidth]{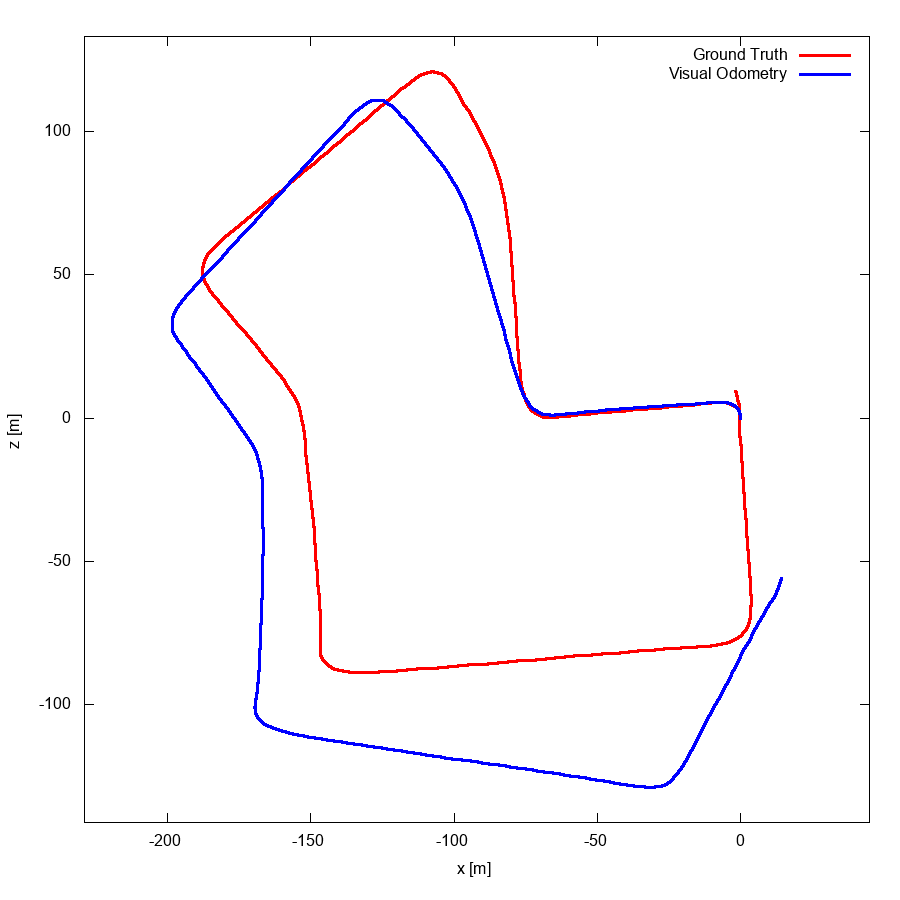} \hfill
 \includegraphics[width=0.45\linewidth]{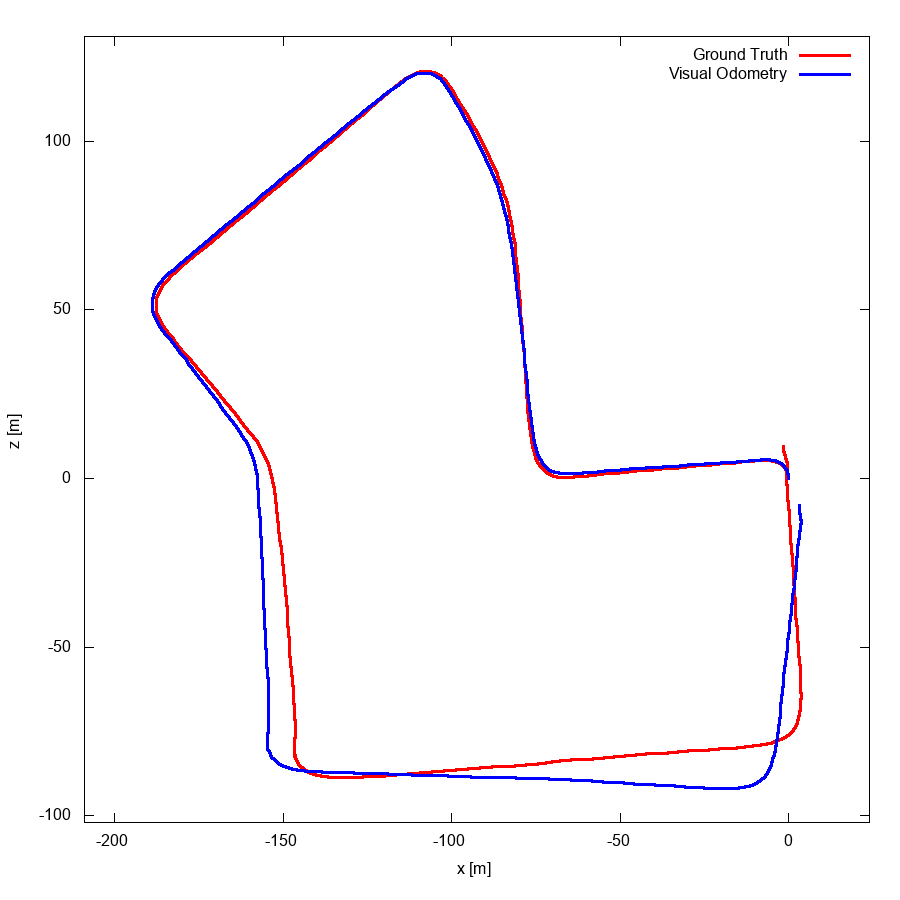}
 \subfigure[unoptimized \DLO{}]{\includegraphics[width=0.45\linewidth]{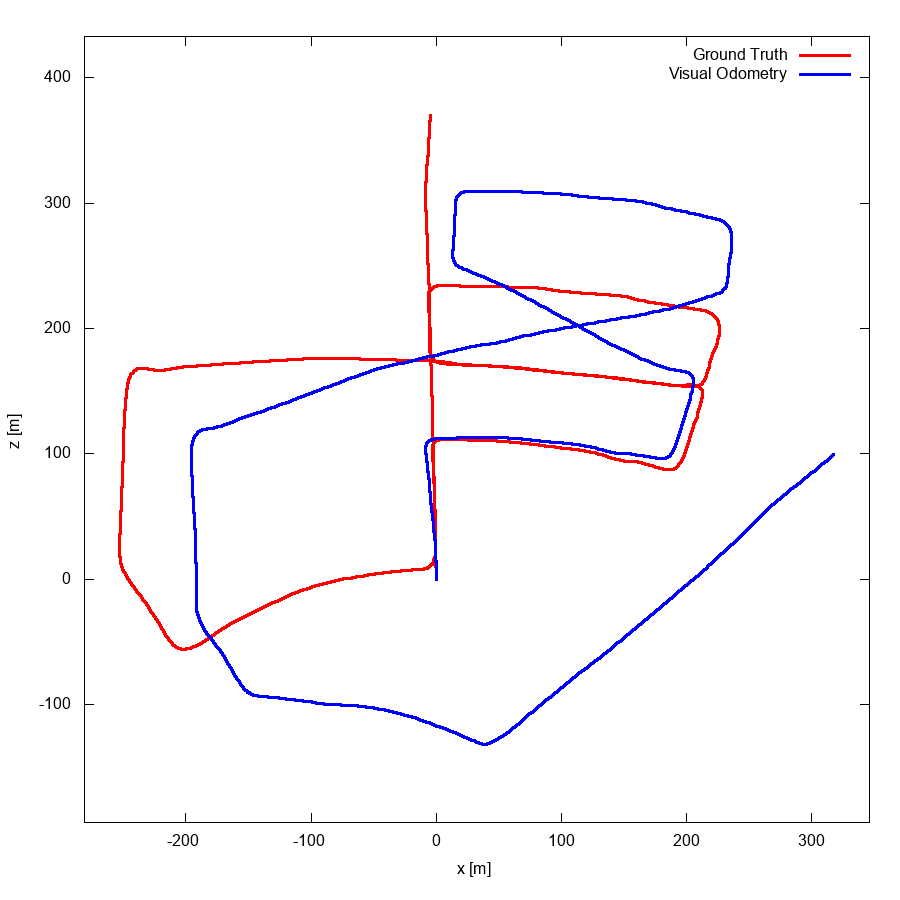}} \hfill
 \subfigure[optimized \DLO{}]{\includegraphics[width=0.45\linewidth]{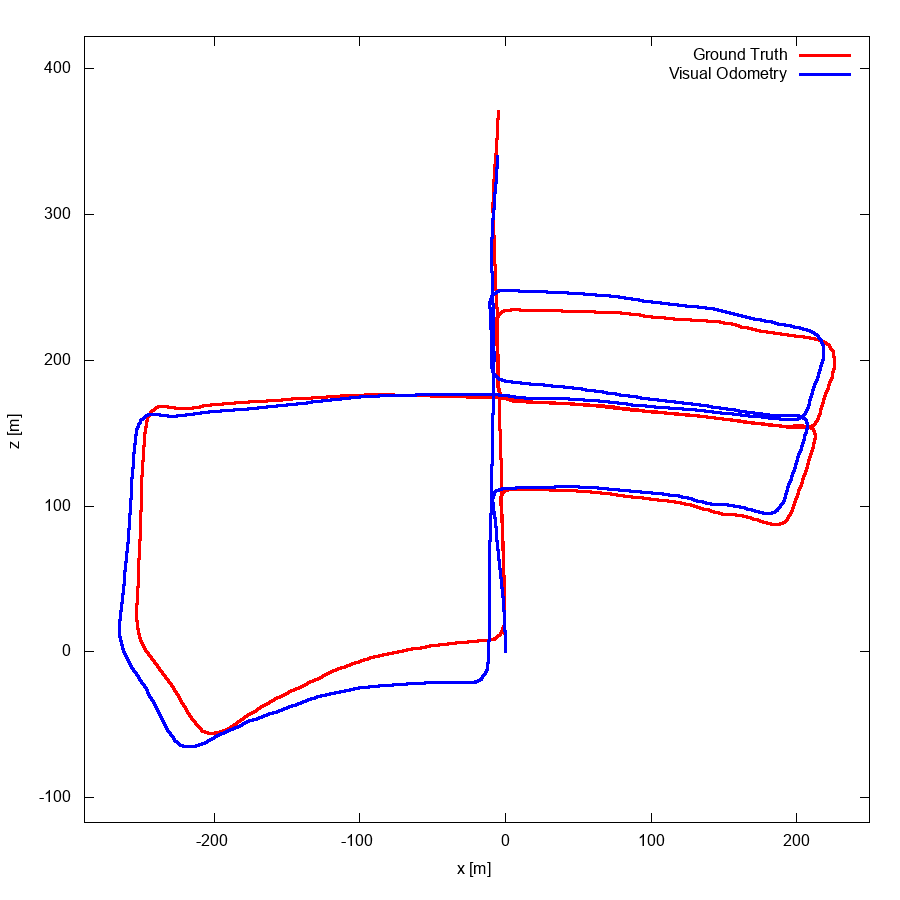}}
\end{center}
   \caption{Estimated trajectories of \textit{training} sequence \#07 (top) and \#05 (bottom) with and without optimization (cp. Section \ref{sec:optimization}). Ground truth is plotted in red and our estimated \DLO{} trajectory in blue. 
   }
\label{fig:eval_u_o}
\end{figure}

For evaluation of our new \DLO{} approach, we use the KITTI \cite{kitti_benchmark} odometry benchmark. 
This dataset consists of 22 sequences out of which 11 sequences have ground truth trajectories available for training. 
The benchmark includes monochromic and color image sequences, GPS-IMU data and laser scanner data collected in both urban and rural areas of City of Karlsruhe, Germany.
For our work we utilize the laser scanner data only; all other data is neglected.
The laser scanner data is represented as a point cloud acquired by a Velodyne HDL-64E LiDAR. 
The evaluation metrics suggested by KITTI compute the rotation and translation errors separately by evaluating errors as a function of the trajectory length and velocity 
.

Final transformation is computed by combining the rotation (see Section \ref{sec:rotation}) and translation (see Section \ref{sec:translation}) as well as an optimization of this transformation (see Section \ref{sec:optimization}).
When concatenating the relative transformations between scans, a long trajectory of the full sequence can be created as exemplary visible in Figure \ref{fig:eval_u_o} and the supplementary video \footnote{\url{https://youtu.be/HzwwxQfpKNo}}.
The estimated transformation can also be applied over time to the point clouds resulting in an environment map. 
Figure \ref{fig:pointcloud} shows such an environmental map formulated for sequence \#07 validating the functionality of \DLO{}.

Figure \ref{fig:eval_u_o} shows two trajectories corresponding to sequences \#07 and \#05 before and after optimization (cp. Section \ref{sec:optimization}). 
Unoptimized trajectory errors are resulting from few erroneous registration caused due to overturning vehicles, pedestrians and other dynamic objects in the scene. 
Lack of characteristic features in the scene can also introduce  errors. 
Since odometry accumulates small transformations over time, few erroneous frames may result in significant drift towards the end of the trajectory.  
This is reduced to a certain extend using a linear motion model and pose graph optimization. 

\begin{figure}
\centering
 \subfigure[BCC \cite{blo}]{\includegraphics[width=0.45\linewidth]{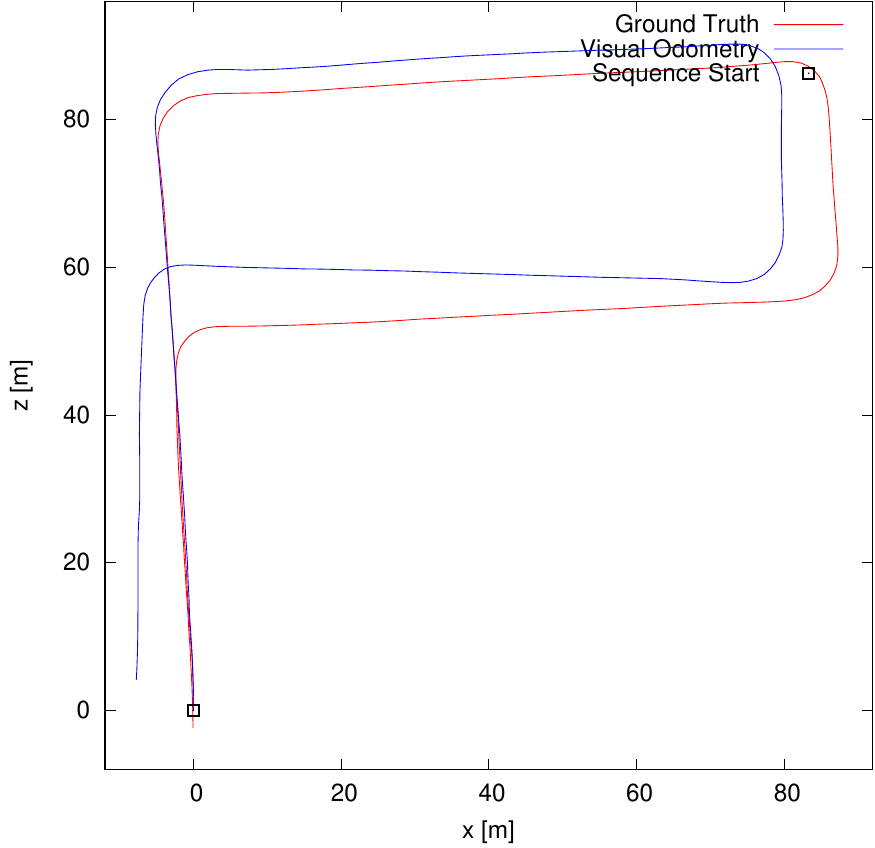}}
 \subfigure[BLO \cite{blo}]{\includegraphics[width=0.45\linewidth]{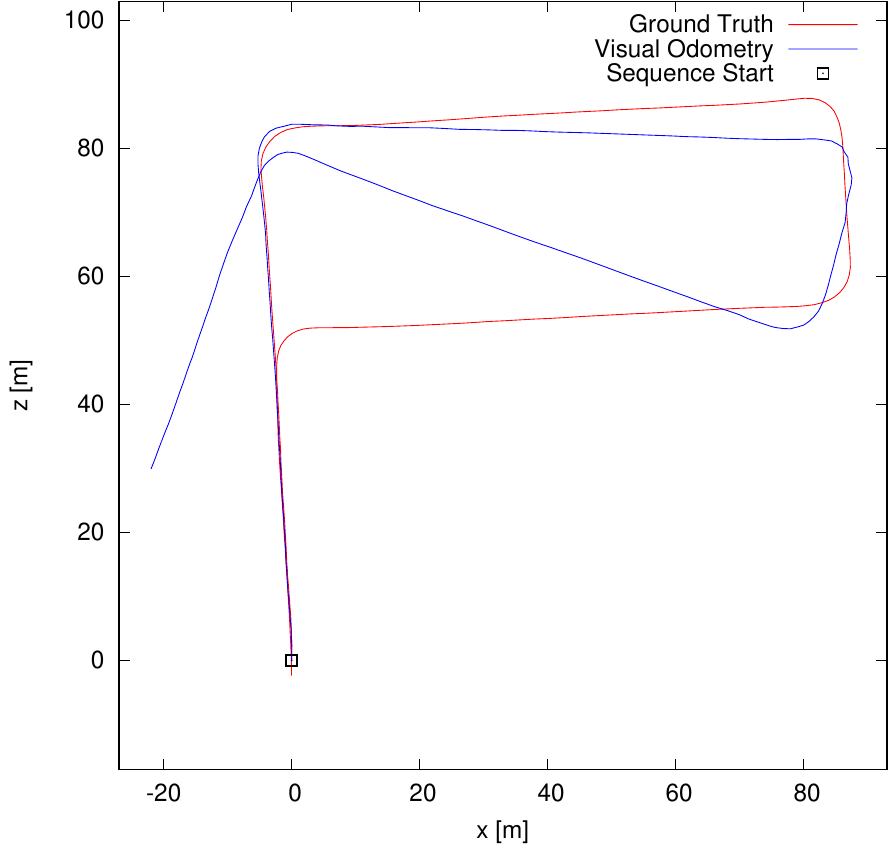}}
 \subfigure[DEMO \cite{zhang2014real}]{\includegraphics[width=0.45\linewidth]{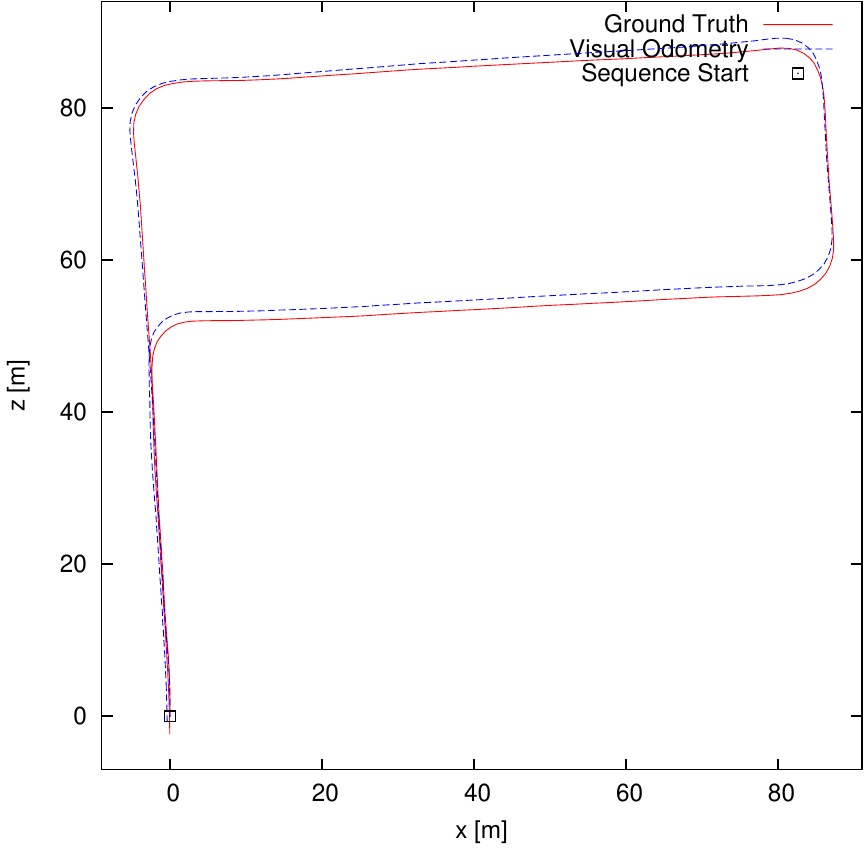}}
 \subfigure[LIMO \cite{graeter2018limo}]{\includegraphics[width=0.45\linewidth]{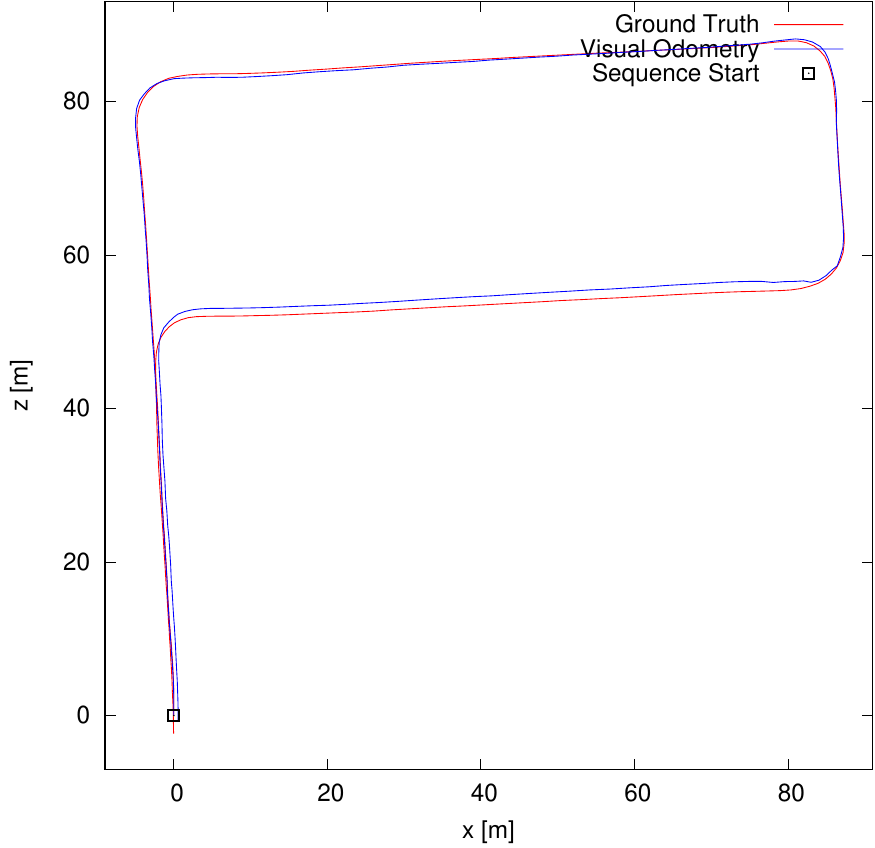}}
 \subfigure[SuMa \cite{behley2018rss}]{\includegraphics[width=0.45\linewidth]{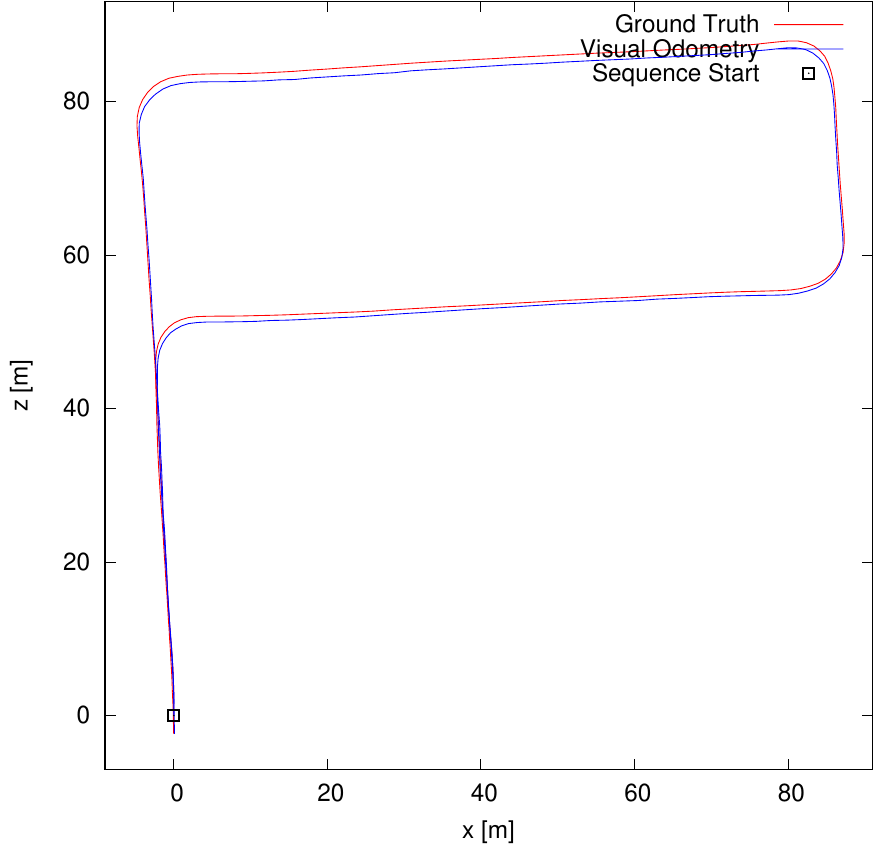}}
 \subfigure[Our \DLO{}]{\includegraphics[width=0.45\linewidth]{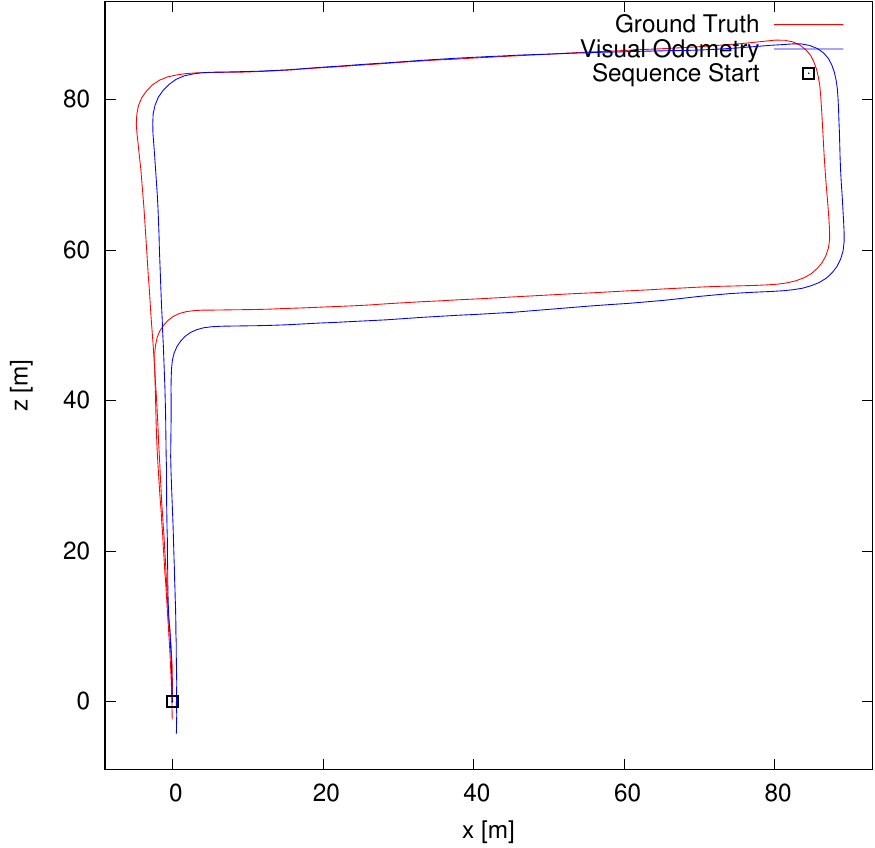}}
 \caption{
   Comparison of our \DLO{} against state-of-the-art approaches on KITTI \textit{test} sequence \#14.
 }
\label{fig:eval_others}
\end{figure}

In Figure \ref{fig:eval_others} the state-of-the-art algorithms are compared against \DLO{} in a ground truth comparison of their trajectories.
While BCC and BLO \cite{blo} show a strong drift, \DLO{} demonstrates similar accuracy to DEMO \cite{zhang2014real}, LIMO \cite{graeter2018limo} and SuMa \cite{behley2018rss}.
This verifies the state-of-the-art performance of our new decoupling approach.

Table \ref{tab:comparison} shows a quantitative comparison of \DLO{} with the state-of-the-art approaches. 
It is important to know that the better performing algorithms utilize additional sensor information or special pre-processing.
For example, LOAM \cite{loam} integrates IMU data for rotation estimation, whereas IMLS \cite{imls} pre-processes the LiDAR scans to remove dynamic objects from the scene. 
LIMO \cite{graeter2018limo} and DEMO \cite{zhang2014real} utilize additional color information.
Our proposed approach \DLO{} uses only LiDAR data without removing the dynamic objects and without using any loop detection.  
The stated error in Table \ref{tab:comparison} results from very few erroneous registrations leading to the increased average value.
Erroneous registration is caused due to overturning vehicles, pedestrians and other dynamic objects in the scene. 
A vehicle moving close to the sensor can result in clusters over the unit sphere. 
Since the algorithm retains only the major clusters/dense regions and rejects everything else from the scene as outliers, the registration may introduce large error and can affect the whole odometry. 
Those frames with no characteristic patterns can also cause substantial errors because of the lack of clusters or reasonable patterns on the unit sphere. 
This is reduced using the optimization techniques. 
Other challenges include relatively high car speed, where the sequences at recorded at 10 fps. 

\begin{table}
\begin{center}
\begin{tabular}{|c|c|c|c|}
\hline
\textbf{Method}&\textbf{Avg trans err}&\textbf{Avg rot err}\\
\hline\hline
LOAM \cite{loam} & 0.59\% & 0.0014[d/m]\\
IMLS \cite{imls}  & 0.64\% & 0.0016[d/m]\\
DEMO \cite{zhang2014real}  & 1.14\% & 0.0049[d/m]\\
SuMA \cite{behley2018rss} & 1.39\% & 0.0034[d/m]\\
BCC \cite{blo} & 4.59\% & 0.0175[d/m]\\
\DLO{} (our) & 5.32\% & 0.0213[d/m]\\
NCICP & 7.17\% & 0.0050[d/m]\\
BLO \cite{blo} & 9.21\% & 0.0163[d/m]\\
\hline
\end{tabular}
\end{center}
\caption{Comparison of \DLO{} with state-of-the-art methods. \DLO{} is showing comparable results using LiDAR scans only. 
}
\label{tab:comparison}
\end{table}


\section{Conclusion}

This paper presents a new algorithm called \DLO{} for odometry using LiDAR point clouds by decoupling rotation from translation. 
The rotation estimation approach based on the distribution pattern of surface normals on unit sphere is independent of translation and thus stands out from the state-of-the-art approaches.
To the best of our knowledge, we are the first presenting such an approach for LiDAR odometry.
In the evaluation we showed the effectiveness of \DLO{} on the well-known KITTI benchmark.
In order to achieve even better results on the benchmark further engineering is necessary.

Consequentially, as future work it will be interesting to remove the dynamic objects from the scenes as proposed in \cite{imls}. 
These dynamic objects are contributing a lot to rotational error.
For the detection of dynamic objects, optical flow \cite{schuster2018flowfields} or scene flow \cite{schuster2018combining} algorithms could be utilized.
Also objects that are observed in one frame and not in another may also be removed to improved the performance. 
Another potential improvement would be to consider a new place recognition algorithm for loop closure like SegMatch \cite{segmatch2017}. 
Detected loop closures could be easily integrated into the already used pose graph and considered for optimization.

\bibliographystyle{IEEEtran}
\bibliography{dlo_bib}

\end{document}